\documentclass[11pt,a4paper]{article}
\usepackage[hyperref]{acl2019}
\usepackage{times}
\usepackage{times}
\usepackage{latexsym}
\usepackage{blindtext}
\usepackage{amsmath}
\usepackage{graphicx}
\usepackage{url}
\usepackage{multirow}
\usepackage{comment}
\usepackage{subcaption}
\usepackage[explicit]{titlesec}
\usepackage{enumitem}
\usepackage{todonotes}

\aclfinalcopy 
\def\aclpaperid{1726} 

\usepackage{booktabs} 
\usepackage{xparse}   
\NewDocumentCommand{\rot}{O{45} O{1em} m}{\makebox[#2][l]{\rotatebox{#1}{#3}}}%

\definecolor{Gray}{gray}{0.5}

\titlespacing{\paragraph}{0em}{
  0.2\baselineskip}{
 0.5 \baselineskip}%
\title{Multi-task Pairwise Neural Ranking for Hashtag Segmentation}

\author{Mounica Maddela\textsuperscript{1}, Wei Xu\textsuperscript{1}, Daniel Preo\c{t}iuc-Pietro\textsuperscript{2} \\ \textsuperscript{1} Department of Computer Science and Engineering, The Ohio State University\\
  \textsuperscript{2} Bloomberg LP \\
  {\tt \{maddela.4, xu.1265\}@osu.edu \quad dpreotiucpie@bloomberg.net}\\
}

\date{}

\begin{document}
\maketitle

\begin{abstract}
Hashtags are often employed on social media and beyond to add metadata to a textual utterance with the goal of increasing discoverability, aiding search, or providing additional semantics. However, the semantic content of hashtags is not straightforward to infer as these represent ad-hoc conventions which frequently include multiple words joined together and can include abbreviations and unorthodox spellings.
We build a dataset of 12,594 hashtags split into individual segments and propose a set of approaches for hashtag segmentation by framing it as a pairwise ranking problem between candidate segmentations.\footnote{Our toolkit along with the code and data are publicly available at \url{https://github.com/mounicam/hashtag_master}}
Our novel neural approaches demonstrate 24.6\% error reduction in hashtag segmentation accuracy compared to the current state-of-the-art method. 
Finally, we demonstrate that a deeper understanding of hashtag semantics obtained through segmentation is useful for downstream applications such as sentiment analysis, for which we achieved a 2.6\% increase in average recall on the SemEval 2017 sentiment analysis dataset.
\end{abstract}

\section{Introduction}

A hashtag is a keyphrase represented as a sequence of alphanumeric characters plus underscore, preceded by the \textit{\#} symbol. Hashtags play a central role in online communication by providing a tool to categorize the millions of posts generated daily on Twitter, Instagram, etc. They are useful in search, tracking content about a certain topic ~\cite{Berardi2011ISTITRECMT,ozdikis2012semantic}, or discovering emerging trends~\cite{sampson2016leveraging}. 


Hashtags often carry very important information, such as emotion~\cite{abdul2017emonet}, sentiment~\cite{saif13}, sarcasm~\cite{bamman2015contextualized}, and named entities \cite{finin2010annotating,ritter2011named}. However, inferring the semantics of hashtags is non-trivial since many hashtags contain multiple tokens joined together, which frequently leads to multiple potential interpretations (e.g.,\ \textit{lion head} vs. \textit{lionhead}). Table~\ref{tbl:examples} shows several examples of single- and multi-token hashtags. While most hashtags represent a mix of standard tokens, named entities and event names are prevalent and pose challenges to both human and automatic comprehension, as these are more likely to be rare tokens. Hashtags also tend to be shorter to allow fast typing, to attract attention or to satisfy length limitations imposed by some social media platforms. Thus, they tend to contain a large number of abbreviations or non-standard spelling variations (e.g., \textit{\#iloveu4eva}) \cite{han2011lexical,eisenstein2013bad}, which hinders their understanding.


\renewcommand{\arraystretch}{1.3}
\begin{table}[t!]
\centering
\small
\begin{tabular}{ lcp{1mm}c}
  \hline
  \multicolumn{1}{c}{\textbf{Type}} & \textbf{Single-token} & &\textbf{Multi-token} \\
  \hline
  Named-entity (33.0\%)&\textit{\#lionhead} & & \textit{\#toyota\textcolor{brown}{\textbf{prius}}} \\ \hline
  Events (14.8\%) & \textit{\#oscars} & &  \textit{\#ipv6\textcolor{brown}{\textbf{summit}}} \\ \hline
  Standard (43.6\%) & \textit{\#snowfall} & & \textit{\#epic\textcolor{brown}{\textbf{fall}}} \\ \hline
  Non-standard (11.2\%) & \textit{\#sayin} & & \textit{\#i\textcolor{brown}{\textbf{love}}u\textcolor{brown}{\textbf{4eva}}} \\\hline
\end{tabular}
\caption{Examples of single- (47.1\%) and multi-word hashtags (52.9\%) and their categorizations based on a sample of our data.}
\label{tbl:examples}
\end{table}

\renewcommand{\arraystretch}{1.0}


The goal of our study is to build efficient methods for automatically splitting a hashtag into a meaningful word sequence. Our contributions are:
\begin{itemize}[noitemsep,topsep=0pt,leftmargin=1em]
    \item A larger and better curated dataset for this task;
    \item Framing the problem as pairwise ranking using novel neural approaches, in contrast to previous work which ignored the relative order of candidate segmentations;
    \item A multi-task learning method that uses different sets of features to handle different types of hashtags;
    \item Experiments demonstrating that hashtag segmentation improves sentiment analysis on a benchmark dataset.
\end{itemize}

Our new dataset includes segmentation for 12,594 unique hashtags and their associated tweets annotated in a multi-step process for higher quality than the previous dataset of 1,108 hashtags~\cite{BansalBV15}. We frame the segmentation task as a pairwise ranking problem, given a set of candidate segmentations. We build several neural architectures using this problem formulation which use corpus-based, linguistic and thesaurus based features. We further propose a multi-task learning approach which jointly learns segment ranking and single- vs. multi-token hashtag classification. The latter leads to an error reduction of 24.6\% over the current state-of-the-art. Finally, we demonstrate the utility of our method by using hashtag segmentation in the downstream task of sentiment analysis. Feeding the automatically segmented hashtags to a state-of-the-art sentiment analysis method on the SemEval 2017 benchmark dataset results in a 2.6\% increase in the official metric for the task.

\section{Background and Preliminaries}
\label{sec:background}

Current approaches for hashtag segmentation can be broadly divided into three categories: (a) gazeteer and rule based \cite{maynard2014cares,declerck2015processing, belainine-fonseca-sadat:2016:WNUT}, (b) word boundary detection \cite{Celebi17, CELEBI16.708}, and (c) ranking with language model and other features \cite{Wang:2011:WSN:1963405.1963457, BansalBV15, Berardi2011ISTITRECMT,reuter2016,Simeon2016}. Hashtag segmentation approaches draw upon work on compound splitting for languages such as German or Finnish~\cite{koehn2003empirical} and word segmentation~\cite{peng2001hierarchical} for languages with no spaces between words such as Chinese~\cite{sproat1990statistical,xue2003chinese}. Similar to our work, \citeauthor{BansalBV15} \shortcite{BansalBV15} extract an initial set of candidate segmentations using a sliding window, then rerank them using a linear regression model trained on lexical, bigram and other corpus-based features. The current state-of-the-art approach \cite{Celebi17,CELEBI16.708} uses maximum entropy and CRF models with a combination of language model and hand-crafted features to predict if each character in the hashtag is the beginning of a new word. 



\noindent \textbf{Generating Candidate Segmentations.} Microsoft Word Breaker \cite{Wang:2011:WSN:1963405.1963457} is, among the existing methods, a strong baseline for hashtag segmentation, as reported in~\citet{Celebi17} and~\citet{BansalBV15}. It employs a beam search algorithm to extract $k$ best segmentations as ranked by the n-gram language model probability:
\begin{equation*}
\begin{aligned}
Score^{LM}(s) & = \sum_{i = 1}^{n} \log P( w_{i} | w_{i - N + 1} \dots w_{i - 1}) 
\end{aligned}
\end{equation*}
where $[w_1, w_2 \dots w_n]$ is the word sequence of segmentation $s$ and $N$ is the window size. More sophisticated ranking strategies, such as Binomial and word length distribution based ranking, did not lead to a further improvement in performance~\cite{Wang:2011:WSN:1963405.1963457}. The original Word Breaker was designed for segmenting URLs using language models trained on web data. In this paper, we reimplemented\footnote{To the best of our knowledge, Microsoft discontinued its Word Breaker and Web Ngram API services in early 2018.} and tailored this approach to segmenting hashtags by using a language model specifically trained on Twitter data (implementation details in \S \ref{sec:implementation}). The performance of this method itself is competitive with state-of-the-art methods (evaluation results in \S \ref{sec:mainresults}). Our proposed pairwise ranking method will effectively take the top $k$ segmentations generated by this baseline as candidates for reranking. 

However, in prior work, the ranking scores of each segmentation were calculated independently, ignoring the relative order among the top $k$ candidate segmentations. To address this limitation, we utilize a
pairwise ranking strategy for the first time for this task and propose neural architectures to model this.



\section{Multi-task Pairwise Neural Ranking}

\begin{figure*}[ht!]
\captionsetup[subfigure]{justification=centering}
\begin{subfigure}[b]{0.22\textwidth}
\centering
\includegraphics[width=0.8in]{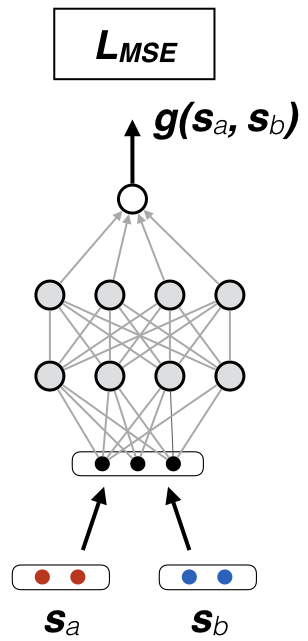} 
\vspace{.19in}
\caption{Pairwise Ranking Model (\textbf{MSE} \S \ref{sec:baseline})}
\end{subfigure}
\begin{subfigure}[b]{0.37\textwidth}
\centering
\includegraphics[width=1.5in]{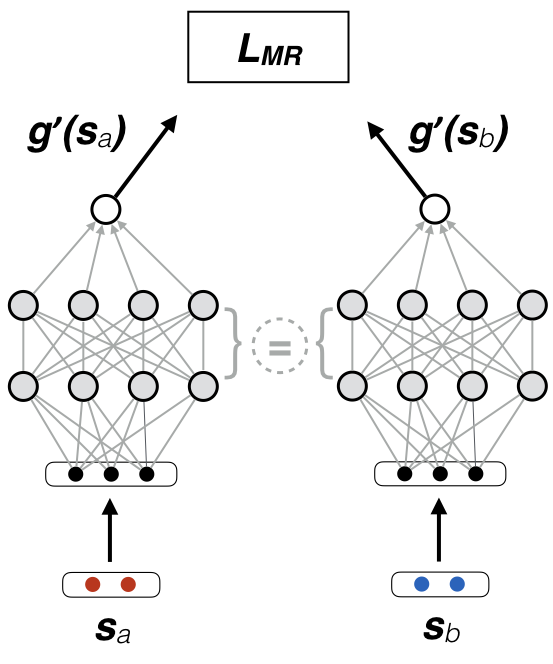}
\vspace{.07in}
\caption{Margin Ranking Loss w/ shared parameters (\textbf{MR} \S \ref{sec:rankloss})}
\end{subfigure}
\begin{subfigure}[b]{0.39\textwidth}
\centering
\includegraphics[width=1.4in]{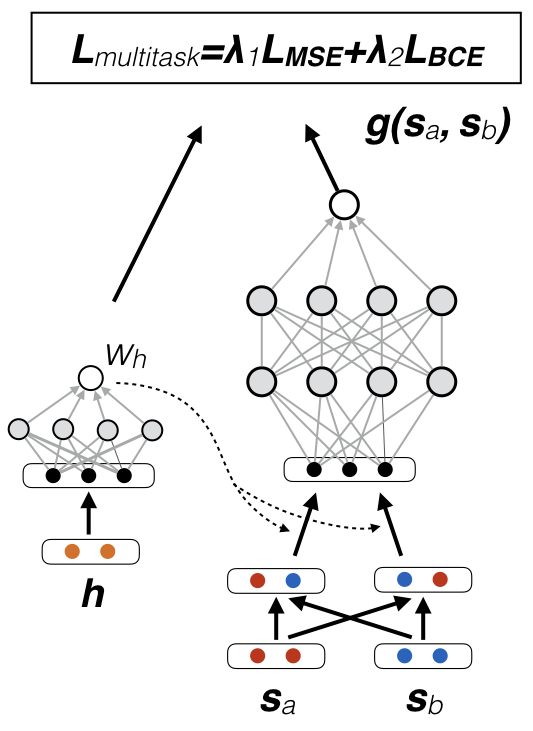}
\vspace{-.05in}
\caption{Adaptive Multi-task Learning for Pairwise ranking (\textbf{MSE+Multitask} \S \ref{sec:multitask})}
\end{subfigure}

\caption{Pairwise neural ranking models for hashtag segmentation. Given two candidate segmentations $s_{a}$ and $s_{b}$ of hashtag $h$, the goal is to predict the segmentation's \textit{goodness} relative score ($g$) or absolute ($g'$) score.}
\label{fig:models}
\end{figure*}

We propose a multi-task pairwise neural ranking approach to better incorporate and distinguish the relative order between the candidate segmentations of a given hashtag. Our model adapts to address single- and multi-token hashtags differently via a multi-task learning strategy without requiring additional annotations. In this section, we describe the task setup and three variants of pairwise neural ranking models (Figure \ref{fig:models}).

\begin{table}[h]
\small
\centering
\begin{tabular}{ll}
\hline
hashtag ($h$) & 
\textit{\#songsonghaddafisitunes}\\ 
\hline
segmentation ($s^{*}$) & \textit{songs} \textit{on} \textit{ghaddafi} \textit{s} \textit{itunes} \\
\multicolumn{2}{r}{(i.e. \textit{songs} \textit{on} \textit{Ghaddafi's} \textit{iTunes})}\\
\hline
\multicolumn{2}{l}{candidate segmentations ($s \in S$)}  \\
& \textit{songs} \textit{on} \textit{ghaddafis} \textit{itunes} \\
& \textit{songs} \textit{on} \textit{ghaddafisi} \textit{tunes} \\
& \textit{songs} \textit{on} \textit{ghaddaf} \textit{is} \textit{itunes} \\
& \textit{song} \textit{song} \textit{haddafis} \textit{i} \textit{tunes} \\
& \textit{songsong} \textit{haddafisitunes} \\
& \multicolumn{1}{r}{(and $\dots$)}  \\
\hline
\end{tabular}
\caption{Example hashtag along with its gold and possible candidate segmentations.}
\label{table:seg_example}
\end{table}

\subsection{Segmentation as Pairwise Ranking}
The goal of hashtag segmentation is to divide a given hashtag $h$ into a sequence of meaningful words $s^{*}=[w_1, w_2, \dots , w_n]$. For a hashtag of $r$ characters, there are a total of $2^{r-1}$ possible segmentations but only one, or occasionally two, of them ($s^{*}$) are considered correct (Table \ref{table:seg_example}).

We transform this task into a pairwise ranking problem: given $k$ candidate segmentations \{$s_{1}, s_{2},\ldots,s_{k}$\}, we rank them by comparing each with the rest in a pairwise manner. More specifically, we train a model to predict a real number $g(s_a, s_b)$ for any two candidate segmentations $s_a$ and $s_b$ of hashtag $h$, which indicates $s_a$ is a better segmentation than $s_b$ if positive, and vice versa. To quantify the quality of a segmentation in training, we define a \textit{gold} scoring function $g^*$ based on the similarities with the ground-truth segmentation $s^{*}$: 
\begin{equation*}
g^*(s_a, s_b) = sim(s_a, s^{*}) - sim(s_b, s^{*}). 
\end{equation*}
We use the Levenshtein distance (minimum number of single-character edits) in this paper, although it is possible to use other similarity measurements as alternatives. We use the top $k$ segmentations generated by Microsoft Word Breaker (\S \ref{sec:background}) as  initial candidates.

\subsection{Pairwise Neural Ranking Model}
\label{sec:baseline}
For an input candidate segmentation pair $\langle s_{a}, s_{b} \rangle$, we concatenate their feature vectors $\mathbf{s}_a$ and $\mathbf{s}_b$, and feed them into a feedforward network which emits a comparison score $g(s_a, s_b)$. The feature vector $\mathbf{s}_a$ or $\mathbf{s}_b$ consists of language model probabilities using Good-Turing \cite{good1953population} and modified Kneser-Ney smoothing \cite{kneser1995improved,chen1999empirical}, lexical and linguistic features (more details in \S \ref{sec:features}). For training, we use all the possible pairs $\langle s_a,s_b \rangle$ of the $k$ candidates as the input and their \textit{gold} scores $g^*(s_a,s_b)$ as the target. The training objective is to minimize the Mean Squared Error (MSE):
\begin{equation}
L_{MSE} = \frac{1}{m} \sum_{i=1}^{m} (g^{*(i)}(s_a,s_b) - \hat{g}^{(i)}(s_a, s_b))^2 
\end{equation}
where $m$ is the number of training examples. 

To aggregate the pairwise comparisons, we follow a greedy algorithm proposed by \citeauthor{cohen1998learning} \shortcite{cohen1998learning} and used for preference ranking \cite{parakhin}. For each segmentation $s$ in the candidate set $S = \{s_1,s_2,\dots,s_k\}$, we calculate a single score $Score^{PNR}(s) = \sum_{s \neq s_{j} \in S} g(s, s_j)$, and find the segmentation $s_{max}$ corresponding to the highest score. We repeat the same procedure after removing $s_{max}$ from $S$, and continue until $S$ reduces to an empty set. Figure~\ref{fig:models}(a) shows the architecture of this model.

\subsection{Margin Ranking (MR) Loss}
\label{sec:rankloss}
As an alternative to the pairwise ranker (\S\ref{sec:baseline}), we propose a pairwise model which learns from candidate pairs $\langle s_a, s_b\rangle$ but ranks each individual candidate directly rather than relatively. We define a new scoring function $g'$ which assigns a higher score to the better candidate, i.e., $g'(s_a) > g'(s_b)$, if $s_a$ is a better candidate than $s_b$  and vice-versa. Instead of concatenating the features vectors $\mathbf{s}_{a}$ and $\mathbf{s}_{b}$, we feed them separately into two identical feedforward networks with shared parameters. During testing, we use only one of the networks to rank the candidates based on the $g'$ scores. For training, we add a ranking layer on top of the networks to measure the violations in the ranking order and minimize the Margin Ranking Loss (MR):
\begin{equation}
\begin{aligned}
L_{MR} & =  \frac{1}{m} \sum_{i=1}^{m} \max(0, 1 - l_{ab}^{(i)} p_{ab}^{(i)}) \\
p_{ab}^{(i)} & = (\hat{g}'^{(i)}(s_a)-\hat{g}'^{(i)}(s_b)) \\
l_{ab} & = \left\{
        \begin{array}{ll}
            1 & \quad g^*(s_a,s_b)  >  0 \\
            -1 & \quad g^*(s_a,s_b) <  0 \\
            0 & \quad \text{otherwise} \\
        \end{array}
    \right.
\end{aligned}
\end{equation}
where $m$ is the number of training samples. The architecture of this model is presented in Figure~\ref{fig:models}(b).

\subsection{Adaptive Multi-task Learning}
\label{sec:multitask}
Both models in \S \ref{sec:baseline} and \S \ref{sec:rankloss} treat all the hashtags uniformly. However, different features address different types of hashtags. By design, the linguistic features capture named entities and multi-word hashtags that exhibit word shape patterns, such as camel case. The ngram probabilities with Good-Turing smoothing gravitate towards multi-word segmentations with known words, as its estimate for unseen ngrams depends on the fraction of ngrams seen once which can be very low \cite{Heafield-thesis}. The modified Kneser-Ney smoothing is more likely to favor segmentations that contain rare words, and single-word segmentations in particular. Please refer to \S \ref{sec:mainresults} for a more detailed quantitative and qualitative analysis.

To leverage this intuition, we introduce a binary classification task to help the model differentiate single-word from multi-word hashtags. The binary classifier takes hashtag features $\mathbf{h}$ as the input and outputs $w_h$, which represents the probability of $h$ being a multi-word hashtag. $w_h$ is used as an adaptive gating value in our multi-task learning setup. The gold labels for this task are obtained at no extra cost by simply verifying whether the ground-truth segmentation has multiple words. We train the pairwise segmentation ranker and the binary single- vs. multi-token hashtag classifier jointly, by minimizing $L_{MSE}$ for the pairwise  ranker and the Binary Cross Entropy Error ($L_{BCE}$) for the classifier:
\begin{equation}
\begin{aligned}
L_{multitask} & =  \lambda_1 L_{MSE} + \lambda_2 L_{BCE} \\
L_{BCE} & =  - \frac{1}{m} \sum_{i=1}^{m} \big[l^{(i)} * log(w_h^{(i)})  + \\
& (1 - l^{(i)}) * log( 1 - w_h^{(i)})\big] \\
\end{aligned}
\end{equation}
where $w_h$ is the adaptive gating value, $l \in \{0,1\}$ indicates if $h$ is actually a multi-word hashtag and $m$ is the number of training examples. $\lambda_1$ and $\lambda_2$ are the weights for each loss. For our experiments, we apply equal weights.

More specifically, we divide the segmentation feature vector $\mathbf{s}_{a}$ into two subsets: (a) $\mathbf{s}_{a}^{KN}$ with modified Kneser-Ney smoothing features, and (b) $\mathbf{s}_{a}^{GL}$ with Good-Turing smoothing and linguistic features. For an input candidate segmentation pair $\langle s_{a}, s_{b} \rangle$, we construct two pairwise vectors $\mathbf{s}_{ab}^{KN}=[\mathbf{s}_{a}^{KN}; \mathbf{s}_{b}^{KN}]$ and $\mathbf{s}_{ab}^{GL}=[\mathbf{s}_{a}^{GL}; \mathbf{s}_{b}^{GL}]$ by concatenation, then combine them based on the adaptive gating value $w_h$ before feeding them into the feedforward network $G$ for pairwise ranking:
\begin{equation}
\hat{g}(s_a, s_b) = G\left(w_h\mathbf{s}_{ab}^{GL} + (1-w_h)\mathbf{s}_{ab}^{KN}\right)
\end{equation} We use summation with padding, as we find this simple ensemble method achieves similar performance  in our experiments as the more complex multi-column networks~\cite{Ciresan12multi-columndeep}. Figure~\ref{fig:models}(c) shows the architecture of this model. An analogue multi-task formulation can also be used for the Margin Ranking loss as:
\begin{equation}
L_{multitask} =  \lambda_1 L_{MR} + \lambda_2 L_{BCE}.
\end{equation}

\subsection{Features}
\label{sec:features}
 
We use a combination of corpus-based and linguistic features to rank the segmentations. For a candidate segmentation $s$, its feature vector $\mathbf{s}$ includes the number of words in the candidate, the length of each word, the proportion of words in an English dictionary\footnote{\url{https://pypi.org/project/pyenchant}} or Urban Dictionary\footnote{\url{https://www.urbandictionary.com}} \cite{nguyen2018emo}, ngram counts from Google Web 1TB corpus \cite{brants2006a}, and ngram probabilities from trigram language models trained on the Gigaword corpus \cite{graff2003} and 1.1 billion English tweets from 2010, respectively. We train two language models on each corpus: one with Good-Turing smoothing using SRILM \cite{Stolcke02srilm--} and the other with modified Kneser-Ney smoothing using KenLM \cite{Heafield:2011:KFS:2132960.2132986}. We also add boolean features, such as if the candidate is a named-entity present in the list of Wikipedia titles, and if the candidate segmentation $s$ and its corresponding hashtag $h$ satisfy certain word-shapes (more details in appendix \ref{sec:rules}). 

Similarly, for hashtag $h$, we extract the feature vector $\mathbf{h}$ consisting of hashtag length, ngram count of the hashtag in Google 1TB corpus \cite{brants2006a}, and boolean features indicating if the hashtag is in an English dictionary or Urban Dictionary, is a named-entity, is in camel case, ends with a number, and has all the letters as consonants. We also include features of the best-ranked candidate by the Word Breaker model. 

\subsection{Implementation Details}
\label{sec:implementation}
We use the PyTorch framework to implement our multi-task pairwise ranking model. The pairwise ranker consists of an input layer, three hidden layers with eight nodes in each layer and hyperbolic tangent ($tanh$) activation, and a single linear output node. The auxiliary classifier consists of an input layer, one hidden layer with eight nodes and one output node with sigmoid activation. We use the Adam algorithm \cite{Kingma2014} for optimization and apply a dropout of 0.5 to prevent overfitting. We set the learning rate to 0.01 and 0.05 for the pairwise ranker and auxiliary classifier respectively. For each experiment, we report results obtained after 100 epochs. 

For the baseline model used to extract the $k$ initial candidates, we reimplementated the Word Breaker \cite{Wang:2011:WSN:1963405.1963457} as described in \S \ref{sec:background} and adapted it to use a language model trained on 1.1 billion  tweets with Good-Turing smoothing using SRILM \cite{Stolcke02srilm--} to give a better performance in segmenting hashtags (\S \ref{sec:mainresults}). For all our experiments, we set $k=10$.

\section{Hashtag Segmentation Data} 
We use two datasets for experiments (Table \ref{table:data_stats}): (a) \textit{STAN$_{small}$}, created by \citeauthor{BansalBV15} \shortcite{BansalBV15}, which consists of 1,108 unique English hashtags from 1,268 randomly selected tweets in the Stanford Sentiment Analysis Dataset \cite{SSAD} along with their crowdsourced segmentations and our additional corrections; and (b) \textit{STAN$_{large}$}, our new expert curated dataset, which includes all 12,594 unique English hashtags and their associated tweets from the same Stanford dataset.

\paragraph{Dataset Analysis.}
\textit{STAN$_{small}$} is the most commonly used dataset in previous work. However, after reexamination, we found annotation errors in 6.8\%\footnote{More specifically, 4.8\% hashtags is missing one of the two acceptable segmentations and another 2.0\% is incorrect segmentation.} of the hashtags in this dataset, which is significant given that the error rate of the state-of-the-art models is only around 10\%. Most of the errors were related to named entities. For example, \textit{\#lionhead}, which refers to the ``Lionhead'' video game company, was labeled as ``\textit{lion head}''.

\begin{table}
\small
\centering
\begin{tabular}{p{1.3cm}cp{2.1cm}cc}
\hline
& \multirow{2}*{\textbf{Data}} & \multicolumn{1}{c}{\textbf{num. of Hashtags}} 
& \textbf{avg.} & \textbf{avg.} \\
& & \multicolumn{1}{r}{\textbf{(multi-token\%)}} & \textbf{\#char}  & \textbf{\#word} \\
\hline
 & Train & \multicolumn{1}{r}{2518 (51.9\%)} & \multicolumn{1}{c}{8.5} & \multicolumn{1}{c}{1.8} \\
\textit{STAN$_{large}$} & Dev & \multicolumn{1}{r}{629 (52.3\%)} & 8.4 & 1.7 \\
 & Test & \multicolumn{1}{r}{9447 (53.0\%)} & 8.6 & 1.8 \\
\hline
\textit{STAN$_{small}$} & Test & \multicolumn{1}{r}{1108 (60.5\%)} & 9.0 & 1.9 \\
\hline
\end{tabular}
\small
\caption{Statistics of the \textit{STAN$_{small}$} and \textit{STAN$_{large}$} datasets -- number of unique hashtags, percentage of multi-token hashtags, average length of hashtags in characters and words.}
\label{table:data_stats}
\end{table}

\paragraph{Our Dataset.}
We therefore constructed the \textit{STAN$_{large}$} dataset of 12,594 hashtags with additional quality control for human annotations. We displayed a tweet with one highlighted hashtag on the Figure-Eight\footnote{\url{https://figure-eight.com}} (previously known as CrowdFlower) crowdsourcing platform and asked two workers to list all the possible segmentations. For quality control on the platform, we displayed a test hashtag in every page along with the other hashtags. If any annotator missed more than 20\% of the test hashtags, then they were not allowed to continue work on the task. For 93.1\% of the hashtags, out of which 46.6\% were single-token, the workers agreed on the same segmentation.  We further asked three in-house annotators (not authors) to cross-check the crowdsourced annotations using a two-step procedure: first, verify if the hashtag is a named entity based on the context of the tweet; then search on Google to find the correct segmentation(s). We also asked the same annotators to fix the errors in \textit{STAN$_{small}$}. The human upperbound of the task is estimated at $\sim$98\% accuracy, where we consider the crowdsourced segmentations (two workers merged) as correct if at least one of them matches with our expert's segmentations. 



\begin{table*}[ht!]
\centering
\small
\begin{tabular}{lrrcccrrcccrcc}
\hline
& \multicolumn{4}{c}{\textbf{All Hashtags}} && \multicolumn{4}{c}{\textbf{Multi-token}}&&
\multicolumn{3}{c}{\textbf{Single-token}}\\
\cline{2-5}  \cline{7-10}   \cline{12-14}
& \textbf{A@1} & \textbf{F$_1$@1} & \textbf{A@2} & \textbf{MRR} &&
 \textbf{A@1} & \textbf{F$_1$@1} & \textbf{A@2} & \textbf{MRR} &&
 \textbf{A@1} & \textbf{A@2} & \textbf{MRR}
\\ \hline
Original hashtag & 51.0 & 51.0 & --  & -- && 19.1  &  19.1 & -- & -- && 100.0 & -- & -- \\
Rule-based \cite{belainine-fonseca-sadat:2016:WNUT} &  58.1 & 63.5 & --  & -- && 57.6 & 66.5 & -- & -- && 58.8 & -- & -- \\
GATE Hashtag Tokenizer (M\&G, \citeyear{maynard2014cares}) & 73.2 & 77.2 & -- & -- && 71.4 & 78.0 & -- & -- && 76.0 & -- & -- \\ 
Viterbi \cite{Berardi2011ISTITRECMT} & 73.4 & 78.5 &  --  &  --  && 74.5 & 83.1 &  --  &  --  && 71.6 &  --  &  --  \\
MaxEnt \cite{Celebi17} & 92.4 & 93.4 & --  & -- && 91.9  & 93.6 & -- & -- && 93.1 & -- & -- \\
\hline
Word Breaker w/ Twitter LM  & 90.8 & 91.7 & 97.4 & 94.5 && 88.5 & 90.0 & 97.8 & 93.7 && 94.3 & 96.8 & 95.7\\
Pairwise linear ranker  & 88.1 & 89.9 & 97.2 & 93.1 && 83.8 & 86.8 & 97.3 & 91.3 && 94.7 & \textbf{97.0} & 95.9 \\
Pairwise neural ranker (MR) & 92.3 & 93.5 & 98.2 & 95.4 && 90.9 & 92.8 & 99.0 & 95.2 && 94.5 & 96.9 & 95.8\\
Pairwise neural ranker (MSE) & 92.5 & 93.7 & 98.2 & 95.5 && 91.2 & 93.1 & 99.0 & 95.4 && 94.5 & \textbf{97.0} & 95.8 \\
Pairwise neural ranker  (MR+multitask) & 93.0 & 94.3 & 97.8 & 95.7 && 91.5 & 93.7 & 98.7 & 95.4 && 95.2 & 96.6 & 96.0  \\
Pairwise neural ranker  (MSE+multitask) & \textbf{94.5} & \textbf{95.2} & \textbf{98.4} & \textbf{96.6} && \textbf{93.9} & \textbf{95.1} & \textbf{99.4} & \textbf{96.8} && \textbf{95.4} & 96.8 & \textbf{96.2} \\
\hline
Human Upperbound &  98.0 & 98.3 &  --  & -- && 97.8 & 98.2 &  --  & -- && 98.4 &  --  & -- \\
\hline
\end{tabular}
\small
\caption{Evaluation results on the corrected version of  \textit{STAN$_{small}$}. For reference, on the original version of \textit{STAN$_{small}$}, the Microsoft Word Breaker API reported an 84.6\% F$_1$ score and an 83.6\% accuracy for the top one output~\cite{Celebi17}, while our best model (MSE+multitask) reported 89.8\% F$_1$ and 91.0\% accuracy.}
\label{table:results}
\end{table*}

\begin{table*}[ht!]
\centering
\small
\begin{tabular}{lcccccrrcccccc}
\hline
& \multicolumn{4}{c}{\textbf{All Hashtags}} && \multicolumn{4}{c}{\textbf{Multi-token}}&&
\multicolumn{3}{c}{\textbf{Single-token}}\\
\cline{2-5}  \cline{7-10}   \cline{12-14}
& \textbf{A@1} & \textbf{F$_1$@1} & \textbf{A@2} & \textbf{MRR} &
& \textbf{A@1} & \textbf{F$_1$@1} & \textbf{A@2} & \textbf{MRR} &
& \textbf{A@1} & \textbf{A@2} & \textbf{MRR}
\\ \hline
Original hashtag & 55.5 & 55.5 & --  & -- && 16.2  &  16.2 & -- & -- && 100.0 & -- & -- \\
Rule-based \cite{belainine-fonseca-sadat:2016:WNUT} &  56.1 & 61.5 & --  & -- && 56.0 & 65.8 & -- & -- && 56.3 & -- & -- \\
Viterbi \cite{Berardi2011ISTITRECMT} & 68.4 & 73.8 &  --   &  --  && 71.2 & 81.5 &   --  &  --  && 65.0 &   --  &  --  \\
GATE Hashtag Tokenizer (M\&G, \citeyear{maynard2014cares}) & 72.4 & 76.1 & -- & -- && 70.0 & 76.8 &-- & -- && 75.3 & -- & -- \\
MaxEnt \cite{Celebi17} & 91.2 & 92.3 & --  & -- && 90.2 & 92.4 & -- & -- && 92.3 & -- & -- \\
\hline
Word Breaker w/ Twitter LM  & 90.1 & 91.0 & 96.6 & 93.9 && 88.5 & 90.0 & 97.0 & 93.4 && 91.9 & 96.2 & 94.4 \\ 
Pairwise linear ranker  & 89.2 & 91.1 & 96.3 & 93.3 && 84.2 & 87.8 & 95.6 & 91.0 && \textbf{94.8} & \textbf{97.0} & \textbf{95.9} \\
Pairwise neural ranker (MR) & 91.3 & 92.6 & 97.2 & 94.6 && 89.9 & 92.4 & 97.5 & 94.3 && 92.8 & 96.8 & 94.9 \\
Pairwise neural ranker (MSE) & 91.3 & 92.6 & 97.0 & 94.5 && 91.0 & 93.6 & 97.7 & 94.9 && 91.5 & 96.2 & 94.1 \\
Pairwise neural ranker  (MR+multitask) & 91.4 & 92.7 & 97.2 & 94.6 && 90.0 & 92.6 & 97.7 & 94.4 && 92.9 & 96.6 & 94.9 \\
Pairwise neural ranker  (MSE+multitask) & \textbf{92.4} & \textbf{93.6} & \textbf{97.3} & \textbf{95.2} && \textbf{91.9} &  \textbf{94.1} & \textbf{98.0} & \textbf{95.4}&& 
93.0 & 96.5 & 94.9 \\
\hline
Human Upperbound &  98.6 & 98.8 &  --  & -- && 98.0 & 98.4 &  --  & -- && 99.2 & -- & -- \\
\hline
\end{tabular}
\small
\caption{Evaluation results on our \textit{STAN$_{large}$} \textit{test} dataset. For single-token hashtags, the token-level F$_1$@1 is equivalent to segmentation-level A@1. For multi-token cases, A@1 and F$_1$@1 for the original hashtag baseline are non-zero because 11.4\% of the hashtags have more than one acceptable segmentations. Our best model (MSE+multitask) shows a statistically significant improvement ($p<0.05$) over the state-of-the-art approach \cite{Celebi17} based on the paired bootstrap test \cite{KirkpatrickBK12}.}
\label{table:our_results}
\end{table*}

\section{Experiments}

In this section, we present experimental results that compare our proposed method with the other state-of-the-art approaches on hashtag segmentation datasets. The next section will show experiments of applying hashtag segmentation to the popular task of sentiment analysis.


\subsection{Existing Methods}

We compare our pairwise neural ranker with the following baseline and state-of-the-art approaches:
\begin{enumerate}[noitemsep,topsep=0pt,leftmargin=1.5em,label=(\alph*)]
    \item The \textbf{original hashtag} as a single token;
    \item A \textbf{rule-based} segmenter, which employs a set of word-shape rules with an English dictionary \cite{belainine-fonseca-sadat:2016:WNUT};
    \item A \textbf{Viterbi} model which uses word frequencies from a book corpus\footnote{Project Gutenberg \url{http://norvig.com/big.txt}}~\cite{Berardi2011ISTITRECMT};
    \item The specially developed \textbf{GATE Hashtag Tokenizer} from the open source toolkit,\footnote{\url{https://gate.ac.uk/}} which combines dictionaries and gazetteers in a Viterbi-like algorithm~\cite{maynard2014cares};
    \item A maximum entropy classifier (\textbf{MaxEnt}) trained on the \textit{STAN$_{large}$} training dataset. It predicts whether a space should be inserted at each position in the hashtag and is the current state-of-the-art ~\cite{Celebi17};
    \item Our reimplementation of the \textbf{Word Breaker} algorithm which uses beam search and a Twitter ngram language model~\cite{Wang:2011:WSN:1963405.1963457};
    \item A \textbf{pairwise linear ranker} which we implemented for comparison purposes with the same features as our neural model, but using perceptron as the underlying classifier ~\cite{Hopkins:2011:TR:2145432.2145575} and minimizing the hinge loss between $g^{*}$ and a scoring function similar to $g'$. It is trained on the \textit{STAN$_{large}$} dataset.


\end{enumerate}

\subsection{Evaluation Metrics}
We evaluate the performance by the top $k$ ($k=1,2$) accuracy (\textbf{A@1}, \textbf{A@2}), average token-level F$_1$ score (\textbf{F$_1$@1}), and mean reciprocal rank (\textbf{MRR}). In particular, the accuracy and MRR are calculated at the segmentation-level, which means that an output segmentation is considered correct if and only if it fully matches the human segmentation. Average token-level F$_1$ score accounts for partially correct segmentation in the multi-token hashtag cases. 

\begin{table}[t!]
\centering
\small
\begin{tabular}{p{2.4cm}cccccccc}
\hline
& \multicolumn{2}{c}{\textbf{Single}} && \multicolumn{2}{c}{\textbf{Multi}} &&
\multicolumn{2}{c}{\textbf{All}}\\
\cline{2-3}  \cline{5-6} \cline{8-9}
& \textbf{A}  & \textbf{MRR} && \textbf{A}  & \textbf{MRR} && \textbf{A}  & \textbf{MRR} \\ \hline
Kneser-Ney & \textbf{95.4} & \textbf{95.7} && 56.0 & 75.3 && 74.9 & 85.1 \\
Good-Turing (GT)  & 91.4 & 93.5 && 85.9 & 91.8 && 88.6 & 92.6\\
Linguistic (Ling) & 89.4 & 91.7 && 71.6 & 82.6 && 80.1 & 87.0\\
GT + Ling &  92.4 & 93.9 && \textbf{86.2} & \textbf{92.3} && \textbf{88.9} & \textbf{92.7}\\ 
\hline
All Features & 91.1 & 93.1 && 89.0 & 93.7 && 90.0 & 93.4 \\
\hline
\end{tabular}
\small
\caption{Evaluation of automatic hashtag segmentation (MSE) with different features on the \textit{STAN$_{large}$} \textit{dev} set. \textbf{A} denotes accuracy@1. While Kneser-Ney features perform well on single-token hashtags, GT+Ling features perform better on multi-token hashtags.}
\label{table:feature_analysis}
\end{table}

\subsection{Results}
\label{sec:mainresults}

Tables~\ref{table:results} and~\ref{table:our_results} show the results on the \textit{STAN$_{small}$} and \textit{STAN$_{large}$} datasets, respectively. All of our pairwise neural rankers are trained on the 2,518 manually segmented hashtags in the training set of \textit{STAN$_{large}$} and perform favorably against other state-of-the-art approaches. Our best model (MSE+multitask) that utilizes different features adaptively via a multi-task learning procedure is shown to perform better than simply combining all the features together (MR and MSE). We highlight the 24.6\% error reduction on \textit{STAN$_{small}$} and 16.5\% on \textit{STAN$_{large}$} of our approach over the previous SOTA \cite{Celebi17} on the \textbf{Multi-token} hashtags, and the importance of having a separate evaluation of multi-word cases as it is trivial to obtain 100\% accuracy for \textbf{Single-token} hashtags. While our hashtag segmentation model is achieving a very high accuracy@2, to be practically useful, it remains a challenge to get the top one predication exactly correct. Some hashtags are very difficult to interpret, e.g., \textit{\#BTV\textbf{\textcolor{brown}{SMB}}} refers to the Social Media Breakfast (SMB) in Burlington, Vermont (BTV).

The improved Word Breaker with our addition of a Twitter-specific language model is a very strong baseline, which echos the findings of the original Word Breaker paper~\cite{Wang:2011:WSN:1963405.1963457} that having a large in-domain language model is extremely helpful for word segmentation tasks. It is worth noting that the other state-of-the-art system \cite{Celebi17} also utilized a 4-gram language model trained on 476 million tweets from 2009.

\renewcommand{\arraystretch}{1.3}
\begin{table}[t!]
\centering
\footnotesize
\begin{tabular}{cccp{1mm}cp{.1in}lp{1mm}}
     \rot[45][1em]{Kneser-Ney}
    & \rot[45][1em]{Good-Turing}
    & \rot[45][1em]{Linguistic} & & \textbf{count} & & \multicolumn{1}{c}{\textbf{Example Hashtags}} & \\
    \hline
    $\circ$ & $\circ$ & $\circ$ & &  31 & & \textit{\#om\textcolor{brown}{\textbf{nom}}nom} \textit{\#BTV\textcolor{brown}{\textbf{SMB}}} & \\
    \hline
    $\bullet$ & $\circ$ & $\circ$ & & 13 & &  \textit{\#commbank}  \textit{\#mamapedia} & \\
    \hline
    $\circ$ & $\bullet$ & $\circ$ & & 38 & & \textit{\#we\textcolor{brown}{\textbf{want}}mcfly}  \textit{\#wine\textcolor{brown}{\textbf{bar}}sf} & \\
    \hline
    $\circ$ & $\circ$ & $\bullet$ & & 24 & & \#cfp\textcolor{brown}{\textbf{09}} \textit{\#Tech\textcolor{brown}{\textbf{Lunch}}South} & \\ 
    \hline
    $\bullet$ & $\bullet$ & $\circ$ & & 44 & & \textit{\#twittographers} \textit{\#bring\textcolor{brown}{\textbf{back}}} & \\
    \hline
    $\bullet$ & $\circ$ & $\bullet$ & & 16 & & \textit{\#iccw} \textit{\#ecom\textcolor{brown}{\textbf{09}}} & \\
    \hline
    $\circ$ & $\bullet$ & $\bullet$ & & 53 & & \textit{\#Lets\textcolor{brown}{\textbf{Go}}Pens}  \textit{\#epic\textcolor{brown}{\textbf{win}}} & \\
    \hline
    $\bullet$ & $\bullet$ & $\bullet$ & & 420 & & \textit{\#prototype} \textit{\#new\textcolor{brown}{\textbf{york}}} & \\
    \hline
\end{tabular}
\caption{Error ($\circ$) and correct ($\bullet$) segmentation analysis of three pairwise ranking models (MSE) trained with different feature sets  Each row corresponds to one area in the Venn diagram; for example, $\circ$$\circ$$\circ$ is the set of hashtags that all three models failed in the \textit{STAN$_{large}$} \textit{dev} data and $\bullet$$\circ$$\circ$ is the set of hashtags that only the model with Kneser-Ney language model features (but not the other two models) segmented correctly. }
\label{table:error_analysis}
\end{table}
\renewcommand{\arraystretch}{1.0}

\subsection{Analysis and Discussion}

\paragraph{Feature Analysis.} To empirically illustrate the effectiveness of different features on different types of hashtags, we show the results for models using individual feature sets in pairwise ranking models (MSE) in Table \ref{table:feature_analysis}. Language models with modified Kneser-Ney smoothing perform best on single-token hashtags, while Good-Turing and Linguistic features work best on multi-token hashtags, confirming our intuition about their usefulness in a multi-task learning approach. Table \ref{table:error_analysis} shows a qualitative analysis with the first column ($\circ$$\circ$$\circ$) indicating which features lead to correct or wrong segmentations, their count in our data and illustrative examples with human segmentation.

\begin{figure}[t!]
  \begin{minipage}{0.55\linewidth}
    \centering
    \includegraphics[width=1.4in, height=1.0in]{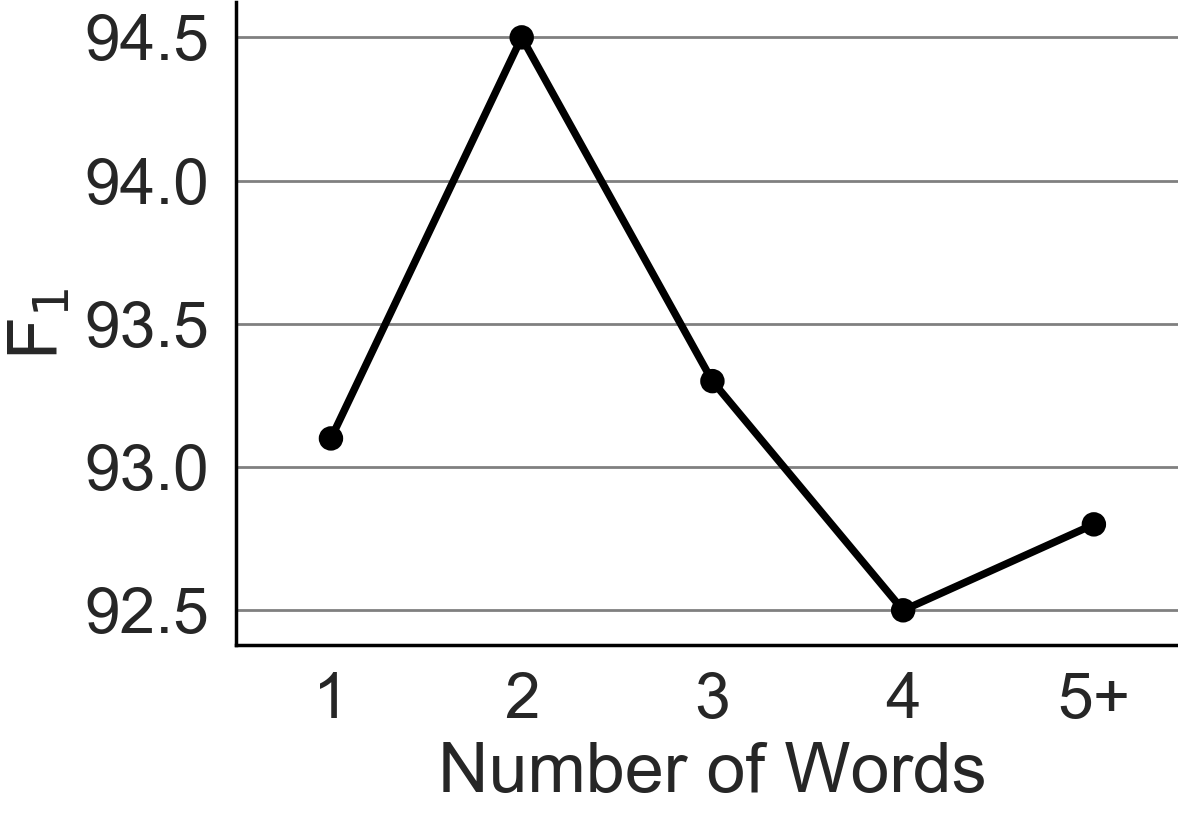}
  \end{minipage}%
  \begin{minipage}[b]{0.35\linewidth}
    \centering
    \scriptsize
\begin{tabular}{ |r|c|}
  \hline
  \multicolumn{1}{|c|}{\textbf{Type}} & \textbf{num. of Hashtags}  \\
  \hline
  single & 4426 (47.1\%) \\ \hline
  2 tokens & 3436 (36.2\%) \\ \hline
  3 tokens & 1085 (11.2\%) \\ \hline
  4 tokens & 279 (2.9\%) \\ \hline
  5+ tokens & 221 (2.6\%) \\ \hline
  \multicolumn{2}{c}{} \\
\end{tabular}
\end{minipage}
\caption{Token-level F$_1$ scores (MSE+multitask) on hashtags of different lengths in the STAN$_{large}$ \textit{test} set. }
\label{fig:perf_analysis}
\end{figure}

\paragraph{Length of Hashtags.} As expected, longer hashtags with more than three tokens pose greater challenges and the segmentation-level accuracy of our best model (MSE+multitask) drops to 82.1\%. For many error cases, our model predicts a close-to-correct segmentation, e.g.,\  \textit{\#you\textbf{\textcolor{brown}{know}}you\textbf{\textcolor{brown}{upt}}too\textbf{\textcolor{brown}{early}}}, \textit{\#isee\textbf{\textcolor{brown}{london}}isee\textbf{\textcolor{brown}{france}}}, which is also reflected by the higher token-level F$_1$ scores across hashtags with different lengths (Figure \ref{fig:perf_analysis}). 

\begin{figure}[t!]
\centering
\includegraphics[width=2.7in]{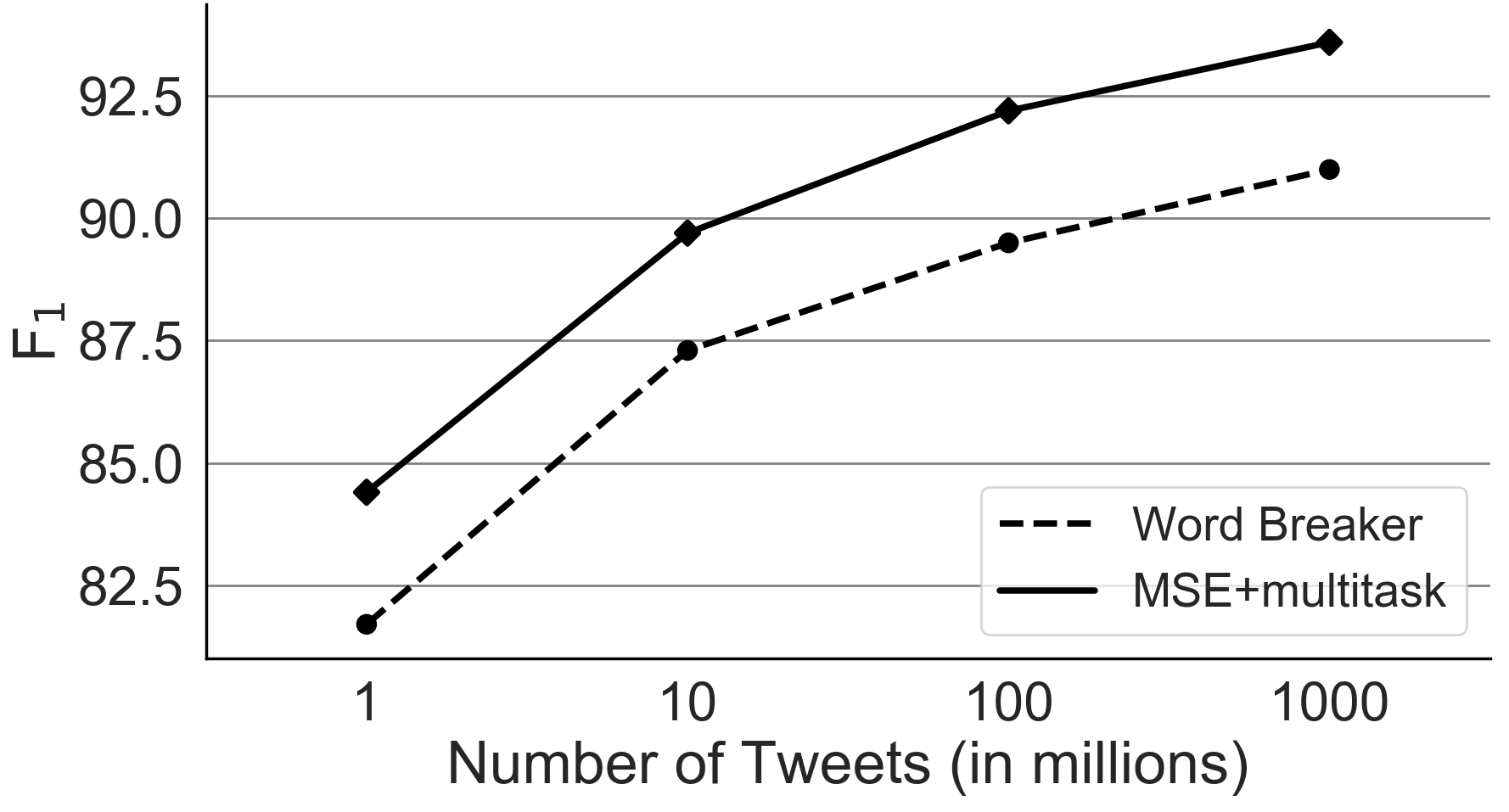}
\caption{Token-level F$_1$ scores of our pairwise ranker (MSE+multitask) and Word Breaker on the STAN$_{large}$ \textit{test} set, using language models trained with varying amounts of data.}
\label{fig:size}
\end{figure}

\paragraph{Size of the Language Model.} Since our approach heavily relies on building a Twitter language model, we experimented with its sizes and show the results in Figure \ref{fig:size}. Our approach can perform well even with access to a smaller amount of tweets. The drop in F$_1$ score for our pairwise neural ranker is only 1.4\% and 3.9\% when using the language models trained on 10\% and 1\% of the total 1.1 billion tweets, respectively. 

\paragraph{Time Sensitivity.} Language use in Twitter changes with time \cite{eisenstein2013bad}. Our pairwise ranker uses language models trained on the tweets from the year 2010. We tested our approach on a set of 500 random English hashtags posted in tweets from the year 2019 and show the results in Table \ref{table:2019}. With a segmentation-level accuracy of 94.6\% and average token-level F$_1$ score of 95.6\%, our approach performs favorably on 2019 hashtags.

\begin{table}[h!]
\small
\centering
\begin{tabular}{p{5.1cm}p{0.7cm}p{0.7cm}p{0.7cm}}
\hline
& \textbf{A@1} & \textbf{F$_1$@1} & \textbf{MRR}
\\ \hline
Word Breaker w/ Twitter LM &  92.1 & 93.9 & 94.7 \\
Pairwise neural ranker (MSE+multitask) &  \textbf{94.6} & \textbf{95.6} & \textbf{96.7} \\
\hline
\end{tabular}
\small
\caption{Evaluation results on 500 random hashtags from the year 2019.} 
\label{table:2019}
\end{table}

\section{Extrinsic Evaluation: Twitter Sentiment Analysis}

We attempt to demonstrate the effectiveness of our hashtag segmentation system by studying its impact on the task of sentiment analysis in Twitter \cite{pang2002thumbs, Nakov2016,rosenthal-farra-nakov:2017:SemEval}. We use our best model (MSE+multitask), under the name \textbf{HashtagMaster}, in the following experiments.

\subsection{Experimental Setup}
We compare the performance of the \textbf{BiLSTM+Lex} \cite{teng-vo-zhang:2016:EMNLP2016} sentiment analysis model under three configurations: (a) tweets with hashtags removed, (b) tweets with hashtags as single tokens excluding the \textit{\#} symbol, and (c) tweets with hashtags as segmented by our system, HashtagMaster. BiLSTM+Lex is a state-of-the-art open source system for predicting tweet-level sentiment \cite{tay-etal-2018-attentive}. It learns a context-sensitive sentiment intensity score by leveraging a Twitter-based sentiment lexicon \cite{tang-EtAl:2014:Coling}. We use the same settings as described by \citeauthor{teng-vo-zhang:2016:EMNLP2016} \shortcite{teng-vo-zhang:2016:EMNLP2016} to train the model. 

We use the dataset from the \textit{Sentiment Analysis in Twitter} shared task (subtask A) at SemEval 2017 \cite{rosenthal-farra-nakov:2017:SemEval}. \footnote{We did not use the Stanford Sentiment Analysis Dataset \cite{SSAD}, which was used to construct the STAN$_{small}$ and STAN$_{large}$ hashtag datasets, because of its noisy sentiment labels obtained using distant supervision.}  Given a tweet, the goal is to predict whether it expresses \textit{POSITIVE}, \textit{NEGATIVE} or \textit{NEUTRAL} sentiment. The training and development sets consist of 49,669 tweets and we use 40,000 for training and the rest for development. There are a total of 12,284 tweets containing 12,128 hashtags in the SemEval 2017 test set, and our hashtag segmenter ended up splitting 6,975 of those hashtags present in 3,384 tweets.

\subsection{Results and Analysis}

\begin{table}[t!]
\small
\centering
\begin{tabular}{lccc}
\hline
& \textbf{AvgR} & \textbf{F$_1^{PN}$} & \textbf{Acc}
\\ \hline
Original tweets &  \textbf{61.7} & 60.0 & 58.7 \\
$-$ No Hashtags &  \textbf{60.2} & 58.8 & 54.2 \\
$+$ Single-word & \textbf{62.3}  & 60.3 & 58.6 \\
$+$ HashtagMaster &  \textbf{64.3} & 62.4 & 58.6 \\
\hline
\end{tabular}
\small
\caption{Sentiment analysis evaluation on the 3384 tweets from SemEval 2017 test set using the BiLSTM+Lex method \cite{tang-EtAl:2014:Coling}. Average recall (AvgR) is the official metric of the SemEval task and is more reliable than accuracy (Acc). F$_1 ^{PN}$ is the average F$_1$ of positive and negative classes. Having the hashtags segmented by our system \textbf{HashtagMaster} (i.e., MSE+multitask) significantly improves the sentiment prediction than not ($p < 0.05$ for AvgR and F$_1 ^{PN}$ against the single-word setup). }
\label{table:tsa}
\end{table}

In Table \ref{table:tsa}, we report the results based on the 3,384 tweets where HashtagMaster predicted a split, as for the rest of tweets in the test set, the hashtag segmenter would neither improve nor worsen the sentiment prediction. Our hashtag segmenter successfully improved the sentiment analysis performance by 2\% on average recall and F$_1^{PN}$ comparing to having hashtags unsegmented. This improvement is seemingly small but decidedly important for tweets where sentiment-related information is embedded in multi-word hashtags and sentiment prediction would be incorrect based only on the text (see Table \ref{table:sentiment_analysis_examples} for examples). In fact, 2,605 out of the 3,384 tweets have multi-word hashtags that contain words in the Twitter-based sentiment lexicon \cite{tang-EtAl:2014:Coling} and 125 tweets contain sentiment words only in the hashtags but not in the rest of the tweet. On the entire test set of 12,284 tweets, the increase in the average recall is 0.5\%.

\section{Other Related Work}

Automatic hashtag segmentation can improve the performance of many applications besides sentiment analysis, such as text classification~\cite{belainine-fonseca-sadat:2016:WNUT}, named entity linking~\cite{BansalBV15} and modeling user interests for recommendations~\cite{chen16}. It can also help in collecting data of higher volume and quality by providing a more nuanced interpretation of its content, as shown for emotion analysis~\cite{qadir2014learning}, sarcasm and irony detection~\cite{maynard2014cares,huang2018disambiguating}. Better semantic analysis of hashtags can also potentially be applied to hashtag annotation \cite{yuewang19}, to improve distant supervision labels in training classifiers for tasks such as sarcasm~\cite{bamman2015contextualized}, sentiment~\cite{saif13}, emotions~\cite{abdul2017emonet}; and, more generally, as labels for pre-training representations of words~\cite{weston2014tagspace}, sentences~\cite{dhingra2016tweet2vec}, and images~\cite{distantimagenet18}.

\renewcommand{\arraystretch}{1.3}
\begin{table}[t!]
\centering
\small
\begin{tabular}{p{1mm}p{2.5in}p{1mm}}
    \hline
    & \textit{\textbf{\textcolor{blue}{Ofcourse}} \#\textbf{\textcolor{red}{clown}}shoes \#alt\textbf{\textcolor{red}{right}} \#Illinois\textbf{\textcolor{red}{Nazis}}} & \\
    \hline
    & \textit{\#\textbf{\textcolor{blue}{Finally}}At\textbf{\textcolor{blue}{peace}}With people calling me ``Kim \textbf{\textcolor{red}{Fatty}} the Third''} & \\
    \hline
    & \textit{Leslie Odom Jr. sang that. \#\textbf{\textcolor{blue}{Thank}}YouObama} &\\
    \hline
    & \textit{After some 4 months of vegetarianism .. it's all the same industry.  \#\textbf{\textcolor{red}{cut}}outthe\textbf{\textcolor{red}{crap}}} & \\
    \hline
\end{tabular}
\caption{Sentiment analysis examples where our HashtagMaster segmentation tool helped. \textcolor{red}{Red} and \textcolor{blue}{blue} words are negative and positive entries in the Twitter sentiment lexicon \cite{tang-EtAl:2014:Coling}, respectively.}
\label{table:sentiment_analysis_examples}
\end{table}

\section{Conclusion}
We  proposed a new pairwise neural ranking model for hashtag segmention and showed significant performance improvements over the state-of-the-art. We also constructed a larger and more curated dataset for analyzing and benchmarking hashtag segmentation methods. We demonstrated that hashtag segmentation helps with downstream tasks such as sentiment analysis. Although we focused on English hashtags, our pairwise ranking approach is language-independent and we intend to extend our toolkit to languages other than English as future work. 




\section*{Acknowledgments}

We thank Ohio Supercomputer Center \cite{Oakley2012} for computing resources and the NVIDIA for providing GPU hardware. We thank Alan Ritter, Quanze Chen, Wang Ling, Pravar Mahajan, and Dushyanta Dhyani for valuable discussions. We also thank the annotators: Sarah Flanagan, Kaushik Mani, and Aswathnarayan Radhakrishnan. This material is based in part on research sponsored by the NSF under grants IIS-1822754 and IIS-1755898, DARPA through the ARO under agreement number W911NF-17-C-0095, through a Figure-Eight (CrowdFlower) AI for Everyone Award and a Criteo Faculty Research Award to Wei Xu. The views and conclusions contained in this publication are those of the authors and should not be interpreted as representing official policies or endorsements of the U.S. Government. 

\bibliography{acl2019}

\begin{thebibliography}{53}
\expandafter\ifx\csname natexlab\endcsname\relax\def\natexlab#1{#1}\fi

\bibitem[{Abdul-Mageed and Ungar(2017)}]{abdul2017emonet}
Muhammad Abdul-Mageed and Lyle Ungar. 2017.
\newblock Emonet: Fine-grained emotion detection with gated recurrent neural
  networks.
\newblock In \emph{Proceedings of the 55th Annual Meeting of the Association
  for Computational Linguistics}, ACL, pages 718--728.

\bibitem[{Bamman and Smith(2015)}]{bamman2015contextualized}
David Bamman and Noah~A Smith. 2015.
\newblock {Contextualized Sarcasm Detection on Twitter}.
\newblock In \emph{Ninth International AAAI Conference on Web and Social
  Media}, ICWSM, pages 574--577.

\bibitem[{Bansal et~al.(2015)Bansal, Bansal, and Varma}]{BansalBV15}
Piyush Bansal, Romil Bansal, and Vasudeva Varma. 2015.
\newblock Towards {D}eep {S}emantic {A}nalysis of {H}ashtags.
\newblock In \emph{Proceedings of the 37th European Conference on Information
  Retrieval}, ECIR, pages 453--464.

\bibitem[{Berardi et~al.(2011)Berardi, Esuli, Marcheggiani, and
  Sebastiani}]{Berardi2011ISTITRECMT}
Giacomo Berardi, Andrea Esuli, Diego Marcheggiani, and Fabrizio Sebastiani.
  2011.
\newblock {ISTI}@{TREC} {M}icroblog {T}rack 2011: {E}xploring the {U}se of
  {H}ashtag {S}egmentation and {T}ext {Q}uality {R}anking.
\newblock In \emph{Text REtrieval Conference (TREC)}.

\bibitem[{Berg{-}Kirkpatrick et~al.(2012)Berg{-}Kirkpatrick, Burkett, and
  Klein}]{KirkpatrickBK12}
Taylor Berg{-}Kirkpatrick, David Burkett, and Dan Klein. 2012.
\newblock An {E}mpirical {I}nvestigation of {S}tatistical {S}ignificance in
  {NLP}.
\newblock In \emph{Proceedings of the 2012 Joint Conference on Empirical
  Methods in Natural Language Processing and Computational Natural Language
  Learning}, EMNLP-CoNLL, pages 995--1005.

\bibitem[{Billal et~al.(2016)Billal, Fonseca, and
  Sadat}]{belainine-fonseca-sadat:2016:WNUT}
Belainine Billal, Alexsandro Fonseca, and Fatiha Sadat. 2016.
\newblock Named {E}ntity {R}ecognition and {H}ashtag {D}ecomposition to
  {I}mprove the {C}lassification of {T}weets.
\newblock In \emph{Proceedings of the 2nd Workshop on Noisy User-generated Text
  (WNUT)}, COLING, pages 102--111.

\bibitem[{Brants and Franz(2006)}]{brants2006a}
Thorsten Brants and Alex Franz. 2006.
\newblock Web {1T} 5-gram {V}ersion 1.
\newblock \emph{Linguistic Data Consortium (LDC)}.

\bibitem[{{\c C}elebi and {\" O}zg{\" u}r(2016)}]{CELEBI16.708}
Arda {\c C}elebi and Arzucan {\" O}zg{\" u}r. 2016.
\newblock Segmenting {H}ashtags using {A}utomatically {C}reated {T}raining
  {D}ata.
\newblock In \emph{Proceedings of the Tenth International Conference on
  Language Resources and Evaluation}, LREC, pages 2981--2985.

\bibitem[{{\c C}elebi and {\" O}zg{\" u}r(2017)}]{Celebi17}
Arda {\c C}elebi and Arzucan {\" O}zg{\" u}r. 2017.
\newblock Segmenting {H}ashtags and {A}nalyzing {T}heir {G}rammatical
  {S}tructure.
\newblock \emph{Journal of Association For Information Science and Technology
  (JASIST)}, 69(5):675--686.

\bibitem[{Center(2012)}]{Oakley2012}
Ohio~Supercomputer Center. 2012.
\newblock Oakley supercomputer.
\newblock \url{http://osc.edu/ark:/19495/hpc0cvqn}.

\bibitem[{Chen and Goodman(1999)}]{chen1999empirical}
Stanley~F Chen and Joshua Goodman. 1999.
\newblock An empirical study of smoothing techniques for language modeling.
\newblock \emph{Computer Speech \& Language}, 13(4):359--394.

\bibitem[{Chen et~al.(2016)Chen, He, and Kan}]{chen16}
Tao Chen, Xiangnan He, and Min-Yen Kan. 2016.
\newblock {Context-aware Image Tweet Modelling and Recommendation}.
\newblock In \emph{Proceedings of the 24th ACM International Conference on
  Multimedia}, MM, pages 1018--1027.

\bibitem[{Ciresan et~al.(2012)Ciresan, Meier, and
  Schmidhuber}]{Ciresan12multi-columndeep}
Dan Ciresan, Ueli Meier, and J\"{u}rgen Schmidhuber. 2012.
\newblock Multi-column {D}eep {N}eural {N}etworks for {I}mage {C}lassification.
\newblock In \emph{Proceedings of the 2012 IEEE Conference on Computer Vision
  and Pattern Recognition}, CVPR, pages 3642--3649.

\bibitem[{Cohen et~al.(1998)Cohen, Schapire, and Singer}]{cohen1998learning}
William~W Cohen, Robert~E Schapire, and Yoram Singer. 1998.
\newblock Learning to {O}rder {T}hings.
\newblock In \emph{Advances in Neural Information Processing Systems}, NIPS,
  pages 451--457.

\bibitem[{Declerck and Lendvai(2015)}]{declerck2015processing}
Thierry Declerck and Piroska Lendvai. 2015.
\newblock Processing and normalizing hashtags.
\newblock In \emph{Proceedings of the International Conference Recent Advances
  in Natural Language Processing}, RANLP, pages 104--109.

\bibitem[{Dhingra et~al.(2016)Dhingra, Zhou, Fitzpatrick, Muehl, and
  Cohen}]{dhingra2016tweet2vec}
Bhuwan Dhingra, Zhong Zhou, Dylan Fitzpatrick, Michael Muehl, and William
  Cohen. 2016.
\newblock {Tweet2Vec: Character-Based Distributed Representations for Social
  Media}.
\newblock In \emph{Proceedings of the 54th Annual Meeting of the Association
  for Computational Linguistics}, ACL, pages 269--274.

\bibitem[{Eisenstein(2013)}]{eisenstein2013bad}
Jacob Eisenstein. 2013.
\newblock {What to do about bad language on the Internet}.
\newblock In \emph{Proceedings of the 2013 Conference of the North American
  Chapter of the Association for Computational Linguistics}, NAACL, pages
  359--369.

\bibitem[{Finin et~al.(2010)Finin, Murnane, Karandikar, Keller, Martineau, and
  Dredze}]{finin2010annotating}
Tim Finin, Will Murnane, Anand Karandikar, Nicholas Keller, Justin Martineau,
  and Mark Dredze. 2010.
\newblock Annotating named entities in {T}witter data with crowdsourcing.
\newblock In \emph{Proceedings of the Workshop on Creating Speech and Language
  Data with Amazon's Mechanical Turk}, NAACL, pages 80--88.

\bibitem[{Go and Huang(2009)}]{SSAD}
Bhayani~R. Go, A. and L.~Huang. 2009.
\newblock Twitter {S}entiment {C}lassification using {D}istant {S}upervision.
\newblock \emph{CS224N Project Report, Stanford}.

\bibitem[{Good(1953)}]{good1953population}
Irving~J Good. 1953.
\newblock The population frequencies of species and the estimation of
  population parameters.
\newblock \emph{Biometrika}, 40(3-4):237--264.

\bibitem[{Graff and Cieri(2003)}]{graff2003}
David Graff and Christopher Cieri. 2003.
\newblock English {G}igaword {LDC2003T05}.
\newblock \emph{Linguistic Data Consortium (LDC)}.

\bibitem[{Han and Baldwin(2011)}]{han2011lexical}
Bo~Han and Timothy Baldwin. 2011.
\newblock Lexical {N}ormalisation of {S}hort {T}ext {M}essages: {M}akn {S}ens
  a\# twitter.
\newblock In \emph{Proceedings of the 49th Annual Meeting of the Association
  for Computational Linguistics}, ACL, pages 368--378.

\bibitem[{Heafield(2011)}]{Heafield:2011:KFS:2132960.2132986}
Kenneth Heafield. 2011.
\newblock {KenLM: Faster and Smaller Language Model Queries}.
\newblock In \emph{Proceedings of the Sixth Workshop on Statistical Machine
  Translation}, WMT, pages 187--197.

\bibitem[{Heafield(2013)}]{Heafield-thesis}
Kenneth Heafield. 2013.
\newblock \emph{Efficient Language Modeling Algorithms with Applications to
  Statistical Machine Translation}.
\newblock Ph.D. thesis, {Carnegie} {Mellon} University.

\bibitem[{Hopkins and May(2011)}]{Hopkins:2011:TR:2145432.2145575}
Mark Hopkins and Jonathan May. 2011.
\newblock Tuning as ranking.
\newblock In \emph{Proceedings of the Conference on Empirical Methods in
  Natural Language Processing}, EMNLP.

\bibitem[{Huang et~al.(2018)Huang, Chen, and Chen}]{huang2018disambiguating}
Hen-Hsen Huang, Chiao-Chen Chen, and Hsin-Hsi Chen. 2018.
\newblock {Disambiguating false-alarm hashtag usages in tweets for irony
  detection}.
\newblock In \emph{Proceedings of the 56th Annual Meeting of the Association
  for Computational Linguistics}, ACL, pages 771--777.

\bibitem[{Kingma and Ba(2014)}]{Kingma2014}
Diederik~P. Kingma and Jimmy Ba. 2014.
\newblock Adam: {A} {M}ethod for {S}tochastic {O}ptimization.
\newblock In \emph{Proceedings of the 3rd International Conference for Learning
  Representations}, ICLR.

\bibitem[{Kneser and Ney(1995)}]{kneser1995improved}
Reinhard Kneser and Hermann Ney. 1995.
\newblock Improved backing-off for m-gram language modeling.
\newblock In \emph{Proceedings of the 1995 International Conference on
  Acoustics, Speech, and Signal Processing}, ICASSP, pages 181--184.

\bibitem[{Koehn and Knight(2003)}]{koehn2003empirical}
Philipp Koehn and Kevin Knight. 2003.
\newblock Empirical methods for compound splitting.
\newblock In \emph{Proceedings of the tenth conference on European chapter of
  the Association for Computational Linguistics}, EACL, pages 187--194.

\bibitem[{Mahajan et~al.(2018)Mahajan, Girshick, Ramanathan, He, Paluri, Li,
  Bharambe, and van~der Maaten}]{distantimagenet18}
Dhruv Mahajan, Ross Girshick, Vignesh Ramanathan, Kaiming He, Manohar Paluri,
  Yixuan Li, Ashwin Bharambe, and Laurens van~der Maaten. 2018.
\newblock {Exploring the Limits of Weakly Supervised Pretraining}.
\newblock In \emph{Tech Report}.

\bibitem[{Maynard and Greenwood(2014)}]{maynard2014cares}
Diana Maynard and Mark~A Greenwood. 2014.
\newblock Who cares about sarcastic tweets? {I}nvestigating the impact of
  sarcasm on sentiment analysis.
\newblock In \emph{Proceedings of the 9th International Conference on Language
  Resources and Evaluation}, LREC, pages 4238--4243.

\bibitem[{Mohammad et~al.(2013)Mohammad, Kiritchenko, and Zhu}]{saif13}
Saif Mohammad, Svetlana Kiritchenko, and Xiaodan Zhu. 2013.
\newblock {NRC-Canada}: Building the state-of-the-art in sentiment analysis of
  tweets.
\newblock In \emph{Proceedings of the Seventh International Workshop on
  Semantic Evaluation}, SemEval, pages 321--327.

\bibitem[{Nakov et~al.(2016)Nakov, Rosenthal, Kiritchenko, Mohammad, Kozareva,
  Ritter, Stoyanov, and Zhu}]{Nakov2016}
Preslav Nakov, Sara Rosenthal, Svetlana Kiritchenko, Saif~M. Mohammad, Zornitsa
  Kozareva, Alan Ritter, Veselin Stoyanov, and Xiaodan Zhu. 2016.
\newblock Developing a successful {S}em{E}val task in sentiment analysis of
  {T}witter and other social media texts.
\newblock \emph{Language Resources and Evaluation}, 50(1):35--65.

\bibitem[{Nguyen et~al.(2018)Nguyen, McGillivray, and Yasseri}]{nguyen2018emo}
Dong Nguyen, Barbara McGillivray, and Taha Yasseri. 2018.
\newblock Emo, love and god: making sense of urban dictionary, a crowd-sourced
  online dictionary.
\newblock \emph{Royal Society Open Science}, 5(5):172320.

\bibitem[{Ozdikis et~al.(2012)Ozdikis, Senkul, and
  Oguztuzun}]{ozdikis2012semantic}
Ozer Ozdikis, Pinar Senkul, and Halit Oguztuzun. 2012.
\newblock Semantic {E}xpansion of {H}ashtags for {E}nhanced {E}vent {D}etection
  in {T}witter.
\newblock In \emph{Proceedings of the 1st international Workshop on Online
  Social Systems}.

\bibitem[{Pang et~al.(2002)Pang, Lee, and Vaithyanathan}]{pang2002thumbs}
Bo~Pang, Lillian Lee, and Shivakumar Vaithyanathan. 2002.
\newblock Thumbs up? {S}entiment {C}lassification using {M}achine {L}earning
  {T}echniques.
\newblock In \emph{Proceedings of the Conference on Empirical Methods in
  Natural Language Processing}, EMNLP, pages 79--86.

\bibitem[{{Parakhin} and {Haluptzok}(2009)}]{parakhin}
M.~{Parakhin} and P.~{Haluptzok}. 2009.
\newblock Finding the {M}ost {P}robable {R}ranking of {O}bjects with
  {P}robabilistic {P}airwise {P}references.
\newblock In \emph{Proceedings of the 10th International Conference on Document
  Analysis and Recognition}, ICDAR, pages 616--620.

\bibitem[{Peng and Schuurmans(2001)}]{peng2001hierarchical}
Fuchun Peng and Dale Schuurmans. 2001.
\newblock A hierarchical em approach to word segmentation.
\newblock In \emph{NLPRS}, pages 475--480.

\bibitem[{Qadir and Riloff(2014)}]{qadir2014learning}
Ashequl Qadir and Ellen Riloff. 2014.
\newblock Learning emotion indicators from tweets: Hashtags, hashtag patterns,
  and phrases.
\newblock In \emph{Proceedings of the 2014 Conference on Empirical Methods in
  Natural Language Processing}, EMNLP, pages 1203--1209.

\bibitem[{Reuter et~al.(2016)Reuter, Pereira-Martins, and Kalita}]{reuter2016}
Jack Reuter, Jhonata Pereira-Martins, and Jugal Kalita. 2016.
\newblock \href {https://doi.org/10.5121/ijnlc.2016.5402} {Segmenting twitter
  hashtags}.
\newblock \emph{International Journal on Natural Language Computing}, 5:23--36.

\bibitem[{Ritter et~al.(2011)Ritter, Clark, Etzioni et~al.}]{ritter2011named}
Alan Ritter, Sam Clark, Oren Etzioni, et~al. 2011.
\newblock {Named Entity Recognition in Tweets: An Experimental Study}.
\newblock In \emph{Proceedings of the Conference on Empirical Methods in
  Natural Language Processing}, EMNLP, pages 1524--1534.

\bibitem[{Rosenthal et~al.(2017)Rosenthal, Farra, and
  Nakov}]{rosenthal-farra-nakov:2017:SemEval}
Sara Rosenthal, Noura Farra, and Preslav Nakov. 2017.
\newblock Sem{E}val-2017 task 4: {S}entiment {A}nalysis in {T}witter.
\newblock In \emph{Proceedings of the 11th International Workshop on Semantic
  Evaluation}, SemEval, pages 502--518.

\bibitem[{Sampson et~al.(2016)Sampson, Morstatter, Wu, and
  Liu}]{sampson2016leveraging}
Justin Sampson, Fred Morstatter, Liang Wu, and Huan Liu. 2016.
\newblock Leveraging the implicit structure within social media for emergent
  rumor detection.
\newblock In \emph{Proceedings of the 25th ACM International on Conference on
  Information and Knowledge Management}, CIKM, pages 2377--2382.

\bibitem[{{Simeon} et~al.(2016){Simeon}, {Hamilton}, and
  {Hilderman}}]{Simeon2016}
C.~{Simeon}, H.~J. {Hamilton}, and R.~J. {Hilderman}. 2016.
\newblock Word segmentation algorithms with lexical resources for hashtag
  classification.
\newblock In \emph{Proceedings of the 2016 IEEE International Conference on
  Data Science and Advanced Analytics (DSAA)}, pages 743--751.

\bibitem[{Sproat and Shih(1990)}]{sproat1990statistical}
Richard Sproat and Chilin Shih. 1990.
\newblock A statistical method for finding word boundaries in chinese text.
\newblock \emph{Computer Processing of Chinese and Oriental Languages},
  4(4):336--351.

\bibitem[{Stolcke(2002)}]{Stolcke02srilm--}
Andreas Stolcke. 2002.
\newblock {SRILM} -- {A}n {E}xtensible {L}anguage {M}odeling {T}oolkit.
\newblock In \emph{Proceedings of the 7th International Conference on Spoken
  Language Processing}, ICSLP, pages 901--904.

\bibitem[{Tang et~al.(2014)Tang, Wei, Qin, Zhou, and
  Liu}]{tang-EtAl:2014:Coling}
Duyu Tang, Furu Wei, Bing Qin, Ming Zhou, and Ting Liu. 2014.
\newblock Building {L}arge-{S}cale {T}witter-{S}pecific {S}entiment {L}exicon :
  {A} {R}epresentation {L}earning {A}pproach.
\newblock In \emph{Proceedings of the 25th International Conference on
  Computational Linguistics}, COLING, pages 172--182.

\bibitem[{Tay et~al.(2018)Tay, Luu, Hui, and Su}]{tay-etal-2018-attentive}
Yi~Tay, Anh~Tuan Luu, Siu~Cheung Hui, and Jian Su. 2018.
\newblock Attentive gated lexicon reader with contrastive contextual
  co-attention for sentiment classification.
\newblock In \emph{Proceedings of the 2018 Conference on Empirical Methods in
  Natural Language Processing}, EMNLP, pages 3443--3453.

\bibitem[{Teng et~al.(2016)Teng, Vo, and Zhang}]{teng-vo-zhang:2016:EMNLP2016}
Zhiyang Teng, Duy~Tin Vo, and Yue Zhang. 2016.
\newblock {Context-Sensitive Lexicon Features for Neural Sentiment Analysis}.
\newblock In \emph{Proceedings of the 2016 Conference on Empirical Methods in
  Natural Language Processing}, EMNLP, pages 1629--1638.

\bibitem[{Wang et~al.(2011)Wang, Thrasher, and
  Hsu}]{Wang:2011:WSN:1963405.1963457}
Kuansan Wang, Christopher Thrasher, and Bo-June~Paul Hsu. 2011.
\newblock {Web Scale NLP: A Case Study on URL Word Breaking}.
\newblock In \emph{Proceedings of the 20th International Conference on World
  Wide Web}, WWW, pages 357--366.

\bibitem[{Wang et~al.(2019)Wang, Li, King, Lyu, and Shi}]{yuewang19}
Yue Wang, Jing Li, Irwin King, Michael~R. Lyu, and Shuming Shi. 2019.
\newblock {Microblog Hashtag Generation via Encoding Conversation Contexts}.
\newblock In \emph{{Proceedings of the North American Chapter of the
  Association for Computational Linguistics (NAACL)}}.

\bibitem[{Weston et~al.(2014)Weston, Chopra, and Adams}]{weston2014tagspace}
Jason Weston, Sumit Chopra, and Keith Adams. 2014.
\newblock \# tagspace: Semantic embeddings from hashtags.
\newblock In \emph{Proceedings of the 2014 Conference on Empirical Methods in
  Natural Language Processing}, EMNLP, pages 1822--1827.

\bibitem[{Xue and Shen(2003)}]{xue2003chinese}
Nianwen Xue and Libin Shen. 2003.
\newblock Chinese word segmentation as {LMR} tagging.
\newblock In \emph{Proceedings of the second SIGHAN workshop on Chinese
  Language Processing}, SIGHAN, pages 176--179.

\end{thebibliography}
\bibliographystyle{acl2019_natbib}

\appendix

\section{Appendix}

\subsection{Word-shape rules}
\label{sec:rules}
Our model uses the following word shape rules as boolean features. If the candidate segmentation $s$ and its corresponding hashtag $h$ satisfies a word shape rule, then the boolean feature is set to True.

\begin{table}[h!]
\centering
\small
\begin{tabular}{r|l}
\hline
\textbf{Rule} & \textbf{Hashtag $\rightarrow$ Segmentation}  \\
\hline
Camel Case & XxxXxx $\rightarrow$ Xxx$+$Xxx  \\
Consonants & cccc $\rightarrow$ cccc \\
Digits as prefix & ddwwww $\rightarrow$ dd$+$wwww  \\
Digits as suffix & wwwwdd $\rightarrow$ wwww$+$dd  \\
Underscore & www$\textunderscore$www $\rightarrow$ www $+$ $\textunderscore$ $+$ www  \\

\hline
\end{tabular}
\small 
\caption{Word-shape rule features used to identify good segmentations. Here, $X$ and $x$ represent capitalized and non-capitalized alphabetic characters respectively, $c$ denotes consonant, $d$ denotes number and $w$ denotes any alphabet or number.}
\label{table:rules}
\end{table}

\end{document}